\documentclass[conference]{IEEEtran}

\usepackage{style/machine-teaching}
\def\BibTeX{{\rm B\kern-.05em{\sc i\kern-.025em b}\kern-.08em
T\kern-.1667em\lower.7ex\hbox{E}\kern-.125emX}}

\begin{document}

\title{Integrating Artificial Intelligence\\into Weapon Systems}

\author{\IEEEauthorblockN{Philip Feldman}
  \IEEEauthorblockA{\textit{ASRC Federal} \\
    Columbia, MD, USA \\
  philip.feldman@asrcfederal.com}
  \and
  \IEEEauthorblockN{Aaron Dant}
  \IEEEauthorblockA{\textit{ASRC Federal} \\
    Columbia, MD, USA \\
  aaron.dant@asrcfederal.com}
  \and
  \IEEEauthorblockN{Aaron Massey}
  \IEEEauthorblockA{\textit{Information Systems} \\
    \textit{University of Maryland, Baltimore County}\\
    Baltimore, MD, USA \\
  akmassey@umbc.edu}
}

\maketitle

\begin{abstract}

  The integration of Artificial Intelligence (AI) into weapon systems is one
  of the most consequential tactical and strategic decisions in the history of
  warfare.  Current AI development is a remarkable combination of accelerating
  capability, hidden decision mechanisms, and decreasing costs.
  Implementation of these systems is in its infancy and exists on a spectrum
  from resilient and flexible to simplistic and brittle.  Resilient systems
  should be able to effectively handle the complexities of a high-dimensional
  battlespace.  Simplistic AI implementations could be manipulated by an
  adversarial AI that identifies and exploits their weaknesses.

  In this paper, we present a framework for understanding the development of
  dynamic AI/ML systems that interactively and continuously adapt to their
  user's needs.  We explore the implications of increasingly capable AI in the
  kill chain and how this will lead inevitably to a fully automated, always
  on system, barring regulation by treaty.  We examine the potential of total
  integration of cyber and physical security and how this likelihood must
  inform the development of AI-enabled systems with respect to the “fog of
  war”, human morals, and ethics.

\end{abstract}

\begin{IEEEkeywords}
  machine learning, computer simulation, human-computer interaction
\end{IEEEkeywords}

\section{Introduction}%
\label{sec:introduction}

John Boyd’s Observation-Orientation-Decision-Action (OODA) model formalizes
the description of the inputs, outputs, experiences, and biases that explain
tactical decision-making for individuals and groups.  In this model, an
adversary attacking either (O)bservation or (O)rientation can create the
conditions for incorrect or catastrophic (D)ecisions and (A)ctions.  Even
without exacerbating conditions such as combat, humans often find
decision-making under stress difficult, particularly with incomplete
information.  The natural human tendency to defer to authorities for
decision-making, human or machine, can also lead to disastrous
outcomes~\cite{ferraris2003nasa, clark2011gps}.

Artificial intelligence and machine learning promise to integrate dynamically
into the human decision-making processes in ways previous technologies could
not, including by being responsive to the operator's cognitive load.  In the
simplest approach, machines perform only tedious and boring tasks.  In a
slightly more complex scenario, machines perform as much of the
non-decision-making activity as possible so that humans can focus completely
on the task at hand.  In the most complex scenario, humans are not directly
involved with the system as it performs the task independently.  These \enquote{always
on} systems respond to threats that are beyond the capability of real-time
human supervision.  Human interaction is restricted to activities such as
training the system in offline or simulated environments

Many countries are currently pursuing an ambitious AI agenda, including the
United States and several potential adversaries~\cite{carter2018chinese,
kania2017battlefield}.  Secretive development of lethal autonomous weapons
systems (LAWS) leads to the conditions for an AI arms race.  Tactically, each
side's human/AI system would be attempting to \enquote{turn inside} the adversaries'
OODA loop.  Although AI may be an advantage in cognitive offloading of mundane
tasks or through increased speed and capability in battle, it presents an
opportunity for a new class of attacks that take advantage of the latent,
high-dimensional spaces in deep neural networks.  These unobserved regions of
the AI decision-making process are prone to \textit{normal accidents} -- a
type of \enquote{inevitable} accident that emerges in situations where components are
densely connected, tightly coupled, and opaque in their
processing~\cite{perrow2011normal}.  Study of this field began with accidents
such as Three-Mile Island, but AI technologies embody similar risks.
Finding and exploiting these weaknesses to induce defective
behavior will become a permanent feature of military
strategy~\cite{staff2014joint}.


This human/AI partnership is likely to produce emergent behaviors that are not
obvious extensions of current military thinking.  This creates a tension
between two poles.  At one end is the need for systems to be trustworthy.
They should predictably do what we believe is the right thing in ethically
difficult conditions.  At the other end is the need to be responsive and
dynamic in unpredictable conditions.  In this paper, we develop a framework
for examining problems in this nascent area of intelligent warfighting machines.

\section{Background}%
\label{sec:background}

Although the battlespace becomes faster and more complex as information
communications technologies improves, the fundamental tactics have been
unchanged for centuries: opposing commanders observe the evolving battlespace,
attempt to understand and model the space, and act to produce positive
outcomes.  Of course, what makes this difficult is that the adversary is doing
the same thing, leading to a co-evolving physical and information environment
that is difficult to predict with any certainty~\cite{dahl1996command}.

These rapidly co-evolving battlespace dynamics are one of the largest
obstacles in involving current state-of-the-art machine learning systems.
Currently, the best AI is based on enormous networks that are trained for days
against massive datasets.  The time frames involved in this process do not
afford the rapid updates that human interaction requires.

To address both the promise and the risks of adding lethal combat capabilities
to AI systems, we need to establish a development framework that emphasizes
human control over the behaviors of such systems, regardless of how
sophisticated they become.  At the core, we believe that these aspects of
human control must include the following:

\begin{description}[leftmargin=1em]

  \item[Interactivity:] Users need to be able to explore and adjust
    the behavior of the system to confirm changes that they made and validate
    that the system exhibits a more “correct” behavior.

  \item[Transparency:] Although intelligent machines may never truly
    be able to explain their actions, they should be able to reveal the
    sources from which they learned any particular behavior.

  \item[Resiliency:] Intelligent systems cannot be brittle.  They must handle
    overload conditions gracefully and recover quickly.  They must be able to
    indicate when they are operating with low confidence, and they cannot
    simply freeze.

\end{description}

\section{Literature Review}%
\label{sec:literature_review}

The goal of adding AI to the battlespace is to augment humans decision-making,
but, adding AI to this decision-making process would have ramifications that
need to be considered carefully.  If AI systems are effective, pressure to
increase the level of assistance to the warfighter would be inevitable.
Continued success would mean gradually pushing the human out of the loop,
first to a supervisory role and then finally to the role of a “killswitch
operator” monitoring an always-on LAWS~\cite{scharre2018army}.

We see four relevant areas of work that address aspects of this problem space:

\begin{description}[leftmargin=1em]

  \item[Cybersecurity:] the virtual counterpart to the physical weapons
    systems

  \item[Hand-to-hand combat:] a proxy for thinking about multiple adversarial
    AI systems of equal capability engaged in combat

  \item[Machine learning research:] how current state-of-the art AI systems
    can responsively and interactively update their states

  \item[Military strategy:] how these systems must operate in the problem
    space

\end{description}

In Section~\ref{sub:cybersecurity}, we examine the current state of the art in
cyberdefense, the limits of a defense-only strategy, and the emerging argument
for cyber-counterattacks, including concerns about automation.  Using that as
a technological frame, we discuss hand-to-hand combat as a proxy for what
happens when there are roughly matched adversarial intelligent systems engaged
in extremely dynamic kinetic actions in Section~\ref{sub:hand_to_hand}.  We
then examine how machine learning research addresses the need for interactive,
evolving dynamic adaptation in Section~\ref{sub:machine_learning}.  Finally,
we fit this information in the frame of military strategy in
Section~\ref{sub:military_strategy}, focusing on the interaction of practical
combat considerations and international law, particularly article 51 of the UN
charter.

\subsection{Cybersecurity and the limits of defense}%
\label{sub:cybersecurity}

Cybersecurity controls and countermeasures often employ several machine
learning and data mining techniques to uncover signs of misuse, anomaly
detection, or hybrid approaches that do
both~\cite{Buczak2016DataMiningSurvey}.  One of the fundamental issues is the
volume and velocity of the information that can be associated with an attack.
Machine learning techniques aid network administrators seeking to respond to
actual issues rather than false alarms, but this approach also represents a
weakness.  Zero day attacks, which have no previously known signature,  can
only be detected using anomaly detection systems.  If successful, a zero day
attack may be able to exploit a system for considerable periods of time.  For
example, the FBI has determined that four individuals with Russian support
were able to penetrate the Yahoo network for two years, getting subscriber
information on 500 million accounts before their activities were detected and
stopped~\cite{fbi_2017}.

Adaptation or generalization from one attack vector to multiple does not
prevent this threat.  The more adaptable the classifier is, the more open it
is to manipulation adversarial training techniques.  In other words, an
adversary can learn the latent spaces in the classifier that lead to false
results.  This can be exploited to overwhelm the system with false positives
for benign vectors while simultaneously rendering dangerous vectors less
detectable~\cite{katzir2018quantifying}.

Academic computer security research is overwhelmingly oriented towards
detecting and blocking cyberattacks.  Regardless of whether the detection
scheme is recognition or anomaly-based, all these approaches rest on the
fundamental assumption that cybersecurity is \textit{passive}.  Systems wait
for attacks, identify them as fast as possible, and determine the best
course(s) of action and respond, often within
milliseconds~\cite{Buczak2016DataMiningSurvey}.

There is a growing awareness among cybersecurity professionals that there may
be a need for active defense as well, particularly if the cyberattack results
in damage to critical national infrastructure.  Active defense may be both
appropriate and effective in eliciting cooperation.  Axelrod showed in 1981
that tit-for-tat responses to aggression were a robust and effective
response~\cite{axelrod1981evolution}.  Two considerations are crucial: The
first is whether counter attack makes sense as a
strategy~\cite{kallberg2017flaw, jensen2002computer}.  The second is how fast
to respond.  Current government processes associated with responding to
kinetic attacks are too slow for responsive
cyberattacks~\cite{grant2017speeding}.

This highlights the fundamental issue in the use of AI systems in weapons
systems, whether virtual or physical.  The feedback loop between
ever-increasing technical capability and the political awareness of the
decreasing time window for reflective decision-making drives technical
evolution towards always-on, automated, reflexive
systems~\cite{scharre2018army}.  This pressure needs to be addressed openly
and transparently in any system design.

\subsection{Hand-to-hand Combat}%
\label{sub:hand_to_hand}

A useful analogy to the evolutionary path that we see happening is individual
unarmed combat.  This is the only example where we can observe a proxy of
similarly matched intelligent systems interacting using the affordances of
force~\cite{miyamoto2002book}.  This model is only effective for considering
evenly matched AI combat systems because in a mismatch, the odds of a rout
are high.  Asymmetric encounters are important as well but not the focus of
this model.

Most human combat consists of two phases: an \textit{assessment phase} where
the adversaries evaluate each other before striking.  This phase is more
analytical and less reflexive.  The adversaries evaluate one another in a
highly dynamic state while moving in tandem.  They employ past training to
generate a plan of action while continuously attempting to lead the opponent
into making incorrect assessments, increasing the size and complexity of the
problem space each opponent has to consider~\cite{hart2009strategy}.  The
second phase is a \textit{kinetic phase} involving rapid strike, defense, and
counterstrike.  For these actions to be effective, they have to be reflexive.
Any reflective thinking slows down the action, exposing vulnerability.

These two processes roughly correspond to Kahneman’s mechanisms for human
cognition~\cite{kahneman2011thinking}.  Kahneman’s System 1 is reflexively
responding to a stimulus, whereas his System 2 is conscious calculation.
System 1 can be \enquote{trained} to respond with seemingly conscious calculation.
A good fighter can produce complex sequences of reflexive action
in response to combat cues.  For example, \textit{The Book of Five
Rings}~\cite{miyamoto2002book}, a canonical work describing traditional
Japanese martial arts, describes the \textit{Crimson-Leaves Strike}, a trained
reflexive action.  The first part of the strike is to identify or cause the
opponent to lower his guard.  This triggers a trained reflex that causes the
fighter to strike reflexively at the opening.

We employ this combat model for the entire range of human and AI combat
systems, from fully human to fully automated.  In all cases, action in the
kinetic phase \textit{must} be as fast as possible, leaving no time to search
for novel solutions.  What changes in the transition to AI systems will be the
speed and number of dimensions to consider.  One can easily imagine
human/computer partnerships, where humans become more involved with the
assessment phase and less with the kinetic phase.  Over time, as AI becomes
more capable of reflective and integrative thinking, the human component will
have to be eliminated altogether as the speed and dimensionality become
incomprehensible, even accounting for cognitive assistance.

\subsection{Machine Learning}%
\label{sub:machine_learning}

Modern machine learning research is focused on developing huge models that
train over even larger datasets, often for days and weeks.  Though startlingly
effective, these systems struggle to adapt to changing conditions
\cite{andersen2018deep}.  One method to increase adaptability is called
\textit{Transfer Learning}~\cite{pan2010survey}, which allows a model
optimized and trained on one dataset to be modified and trained on a
different, smaller but related dataset.  For example, image recognition
systems trained to recognize vehicles for a self-driving car application could
be adapted to detect and recognize military aircraft.

\begin{figure}[hbtp]
  \centering
  \fbox{\includegraphics[width=\columnwidth]{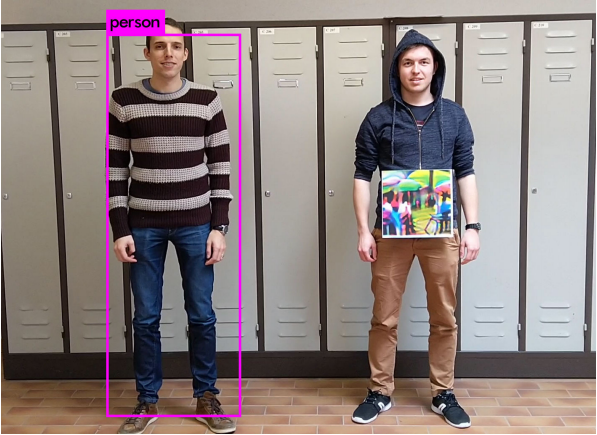}}
  \caption{Adversarial Attack Against an Image Classifier~\cite{thys2019fooling}}
  \label{fig:not_person}
\end{figure}

These kinds of machine learning models contain a weakness.  Numerous studies
have shown that \textit{adversarial attacks} can cause systems to misclassify
examples that are only slightly different from correctly classified
examples”~\cite{goodfellow2014explaining}.  For example,
Figure~\ref{fig:not_person} shows that wearing a picture can fool an image
classifier~\cite{thys2019fooling}.

For combat systems, huge models create an inherent risk.  Because academic and
corporate models are few and often in the public domain, a malicious actor
seeking to save development costs can simply download and study them for areas
where they can be manipulated to respond incorrectly to a particular set of
stimuli~\cite{goodfellow2014explaining}.  However, transfer learning changes
only a small part of the network, latent space vulnerabilities would likely
exist regardless of how the model has been adapted.  This places any combat
system based on one of these models at risk for undetectable exploitation.
Models developed in secure environments based on real and simulated data may
be significantly more secure from exploitation, but with a correspondingly
higher cost to develop.

Neural network systems can learn as they interact with an environment, through
\textit{Reinforcement Learning} (RL)~\cite{suay2011effect}.  This techniques
allows a system to explore a problem space with respect to an evaluation
function that can score the system's behavior.  Such systems can learn to move
in simulated environments, play games, and operate robots.  Adversarial
versions, where one RL system is scored by how well it is competing against
another, currently form the basis for state of the art results in such games
as Chess and Alpha Go.  Training time for adversarial gamespaces is
significant.  DeepMind's AlphaGo Zero took approximately 40 days to train,
including self-play of 29 million games~\cite{silver2017mastering}.  If a less
exhaustive exploration of the data space is acceptable, RL systems can be
guided through their learning process by humans.  This technique significantly
reduces learning time because humans can often recognize incorrect behavior
long before the machine can~\cite{fridman2018human}.



\subsection{Military Strategy}%
\label{sub:military_strategy}

The role of military systems varies by context and can be deeply complicated.
Broadly, forces can be in a peacetime state, at the boundary between peace and
war, and in armed conflict~\cite{sklerov2009solving}.  Article 51 of the UN
Charter specifies that member states may always act in self defense, but there
are now decades of precedent that specify how that right may be interpreted.
Further, as we have seen in the cyberdefense section, the definition of what
constitutes force is changing.

Understanding military contexts matters because of how we develop trust in the
automated systems we use regularly.  Continuous interaction builds trust
incrementally until we implicitly hand off responsibility to the system and
direct our attention elsewhere~\cite{haciyakupoglu2015social,
mosier1996automation}.  This user bias is recognized as the root cause in
numerous accidents.  If your early warning system alerts for an incoming
attack, the pressure to trust the alert and respond is powerful.  System trust
has been a contributing factor in fratricidal battlefield losses involving
Patriot Missiles in Iraq~\cite{hawley2011not}.  Once any level of handoff
occurs, the transfer of control should generally assumed to be total.

When a soldier or commander makes a judgment call about the use of force, an
explicit set of procedures, orders, or other judgment calls precede the first
shot being fired.  This chain of accountability reduces the dimensionality of
the problem space of the decision.  But intelligent machines do not proceed
similarly.  Once a higher-level decision is made, an AI weapon system can
effect orders almost immediately, which is part of the attraction of
integrating AI into conflict operations.

When we start to include AI in weapons systems at any level, the system will
need to know the context it is acting in and the rules of engagement for that
context.  Designers cannot foresee all potential contexts, so the system will
need to be adaptable and transparent.  A human-intensive example of this
approach is used in the Aegis combat system, where the parameters for
semiautomated and fully automated behavior of the system is prepared in
advance of each deployment by the ship's doctrine review board, a diverse mix
of officers and senior enlisted crew that review all aspects of the upcoming
mission before preparing a suite of behavior \enquote{packages} that can activated by
the captain~\cite{scharre2018army}.

Context is critical.  Training exercises may look like war, but they are
actually between allies.  A cold war may look like peace, but it isn't
exactly.  Any intelligent system (human, human/machine, or machine) must be
aware of these and other complicating concerns.  A weapons system that either
directly or indirectly starts a conflict will need to act in accordance with
international law or risk implicating its makers and the government that
activated it in war crimes.  Adversaries know this and will try to use that
weakness against any intelligent system.

The U.S.\ Military does not have and significant technological advantage in
this space.  China, to take one example, views the U.S.\ as highly vulnerable
in cyberwar and is working to cement its potential advantage.  Pressure to
develop systems that can effectively grapple with our adversaries across
multiple domains, dimensions, and timeframes will be extremely high.  All
sides are equally pressured to gain superiority, and as such the inevitability
of fully automated, always on systems should be seriously considered in all
aspects of AI integration.

\section{Opportunities and Challenges}%
\label{sec:opportunities_and_challenges}

Military adaptation of commercially or academically trained models contains
inherent risks.  However, these risks highlight potential opportunities for
development that are distinct from the focus of commercial and academic
communities.  In particular, we identify the following opportunities and
challenges:
\begin{enumerate*}[label=(\Alph*)]

  \item Offline latent space hacking by adversaries;

  \item Incorporating legal and ethical constraints into training a model;

  \item Mapping, traceability, and transparency of inputs and outputs; and

  \item Avoiding dangerous predictability.

\end{enumerate*}

\subsection{Offline Latent Space Hacking by Adversaries}%
\label{sub:latent_space_hacking}

Learning to exploit regions in the available latent space of large models
should be explored in depth.  Less well supported actors may take advantage of
commercial or academic models in an attempt to gain high military impact for
low cost and effort.  Determining how to thwart, for example, a terrorist
organization turning a facial recognition model into a targeting system for
exploding drones is certainly a prudent move.

Technologies such as evolutionary development of model structures in a
reinforcement learning environment can create a framework to support the
development of unique network structures that cannot be predicted by an
adversary~\cite{angeline1994evolutionary}.  Further, such a framework can be
induced to create different networks that address the same sets of problems,
making it possible to generate a diverse set of systems providing redundancy
and resilience.

\subsection{Incorporating Legal and Ethical Constraints}%
\label{sub:legal_and_ethical_constraints}

Modern AI/ML systems reflect the data used to train them.  In commercial and
academic systems, data often reflects unconscious biases that emerge in the
trained behavior of the system~\cite{bolukbasi2016man}.  This type of behavior
only becomes more dangerous when it is connected with weapon systems.  We need
to develop techniques that allow us to train models that have the appropriate
doctrine \enquote{baked in} so that they can operate appropriately in contexts
ranging from war games in peacetime to escorting an adversarial emissary to a
peace conference in wartime.  Although some research exists for encoding legal
and ethical considerations (e.g., using evolutionary
approaches~\cite{honarvar2009artificial}), publications in this area are rare.

\subsection{Mapping, Traceability, and Transparency}%
\label{sub:mapping_traceability_and_transparency}

It is our strong belief that intelligent weapons systems of the future will
move and think at machine speed.  This disproportionate capability and the
inevitable system trust human operators will place in these machines means
that most if not all lethal and sub-lethal interactions will only be
analyzable in hindsight~\cite{carter2019activation}.  Military weapon system models
must be built to support a recorded mapping of inputs that can be traced to
actions or recommendations.  This level of transparency is crucial for
post-incident analysis, validation, and retraining.

\subsection{Lack of diversity}%
\label{sub:lack_of_diversity}

Because the creation of models is complicated and time consuming, few
commercial models address substantively similar tasks at the same level of
sophistication.  Indeed, there is often only one \enquote{best} model for any
set of data.  In either a cybersecurity or military context, this sort of
monoculture represents a predictable single point of failure.  Diverse models
need to be developed to address tasks redundantly, and they must be regularly
revisited, modified, or rebuilt to ensure any adversary obtaining a system
using one model will not be able to rely completely on it in the face of
battle.  This challenge will likely not be met in the academic or commercial
community, where raw performance improvements determines success, not
survivability and ruggedness.

\section{The Role of Humans}%
\label{sec:the_role_of_humans}

Because of the rate of technological development in the AI/ML space, we
believe that the role of humans in combat systems, \textit{barring regulation
  through treaty},\footnote{And assuming that AI/ML systems are not advanced
  enough to autonomously incorporate the risk and consequences of violating
international treaty into their decision-making.} will become more peripheral
over time.  As such, it is critical to ensure that our design decisions and
the implementations of these designs incorporate the values that we wish to
express as a national and global culture.

\subsection{Human-in-the-loop}%
\label{sub:human_in_the_loop}

The starting point for many intelligent systems begins with the tight
integration of human and machine in the weapon system.  For example, missiles
announce when they have a lock, increasing the capability of the warfighter
and leave little ambiguity as to what the weapons system will do once the
trigger is pulled.  If fault has to be found, it will be the human that must
bear the responsibility.

But in more ambiguous circumstance, such as friend-or-foe identification
(IFF), the data used to make the calculation will be in the possession of the
system, not the user.  Imagine a case where an IFF transponders
have been known to be spoofed by the adversary, and a large, slow moving
aircraft identifying as civilian has been detected  on
what seems to be a hostile approach, and only a short time to make a decision.
There are four presentations:
\begin{enumerate*}[label=(\arabic*)]
  \item the system can declare that it has identified the aircraft as hostile
    and provide a lock;
  \item the system can declare the aircraft as friendly and open a channel to
    warn;
  \item the system can present a set of ranked recommendations and provide a
    set of options to the user; or
  \item the system simply displays the raw information.
\end{enumerate*}

The third option may seem to be the best embodiment of the human-in-the-loop
philosophy, but it discounts the effects of system trust.  The user may spend
some time evaluating the list of options the first few times, but if the
system places the correct option at the top of the list often enough, the
human user will begin to simply select that option.  This effect is
exacerbated with time pressures.  The human simply becomes a rubber stamp.

If, however, the system is trained by a set of known individuals, and the
provenance of the system ranking can be traced back to its \enquote{mentor.}
Mentors could train the model in the context of the current deployment and be
known to the user.  The machine then incorporates rules of engagement that are
related to the particular deployment through this human interaction.  Further,
weights that are accumulated from these interactions can be brought back and
integrated into the models, allowing them to evolve with respect to the
current realities of a given battlespace.

\subsection{Human-on-the-loop}%
\label{sub:human-on-the-loop}

As human-in-the-loop systems advance, the system with less need to rely on
human decision-making to achieve results will begin to dominate.  Thus, humans
will be relegated to offline analysis and improvement of AI strategies during
training.  Work is already being done in this space commercially.  Examples of
what is essentially human-on-the-loop architectures are regularly explored now
in StarCraft competitions~\cite{andersen2018deep}.  From a machine learning
perspective, the difference between a StarCraft 2 environment an autonomous
Aegis battlegroup is one of scale and consequences.  Though, human-on-the-loop
systems have less interaction in real-time, integration of the mentor
architecture described above may be possible.  This integration would depend
on the creation of offline wargames and simulations that can be played at
rates slow enough to elicit meaningful training from expert human cognition.

Continually running human-led scenarios offline increases the odds that the
trained system reflects the realities of the
battlespace~\cite{scharre2018army}.  Separate classifier systems may also be
able to catalog adversary behaviors near the boundaries of the current trained
responses.  These boundaries might be detected by looking at the behavior of
the systems themselves and recognizing when they are making decisions among
multiple options with divergent potential outcomes.

\subsection{Human initiated}%
\label{sub:human_initiated}

An extension of the human-on-the-loop approach is the human-initiated “fire
and forget” approach to battlefield AI.  Once the velocity and dimensionality
of the battlespace increase beyond human comprehension, human involvement will
be limited to choosing the temporal and physical bounds of behavior desired in
an anticipated context.  Depending on the immediacy or unpredictability of the
threat, engaging the system manually at the onset of hostilities may be
impossible.  Rather, these systems would need to be activated before the onset
of hostilities.  Activation alone could be an extremely consequential.  Once
activated, shutting the system down for any reason may be interpreted by an
adversarial AI as a weakness to be exploited.  Predicting with confidence how
a conflict between two equally matched, highly capable AI systems would unfold
once started may be impossible.  These conflicts would have to be simulated
extensively against a wide variety of adversaries to have any confidence that
their behavior would align with our national values.

\subsection{Post-hoc forensics}%
\label{sub:posthoc_forensics}

Given a battlespace so overwhelming that humans cannot manually engage with
the system, the human role will be limited to post-hoc forensic analysis, once
hostilities have ceased, or treaties have been signed and
implemented.\footnote{Diplomacy in such an environment seems like a nearly
incomprehensible challenge and deserves extensive research on its own.}  To
this end, these systems will need the maximal amount of provenance of input
data, alternatives considered, and actions selected with sufficiently
high-fidelity recordings that any error or inexplicable behavior can be
meaningfully interpreted~\cite{liu2018interpret}.  Recent work that applies an
approach of this sort is the concept of an \enquote{activation
atlas}~\cite{carter2019activation} that can show areas of conceptual conflict
within a model.  These approaches are nascent, but need to be vigorously
explored and developed.

\section{Discussion and Summary}%
\label{sec:discussion_and_summary}

Tight integration of AI into the kill chain is not a decision to be taken
lightly.  Particularly with respect to LAWS, our policy should be: \textit{not
until they can outperform human/MI collaboration, including making ethically
acceptable choices}~\cite{carter2018chinese}.  Prior to this, deepening our
understanding of the roles, risks, and rewards of integrating AI into the
killchain can better define the problem space of the solutions that we
contemplate.

Many of the solutions needed to address concerns identified herein are likely
also useful in commercial or civilian settings.  The dual use applications for
responsive, diverse and adaptable AI seem numerous, ranging from near-term
problems such as autonomous vehicles in varied environments, to locally
personalized, private, and secure AI assistants.  Novel solutions in these
area could be useful in many sectors of today's economy, and might kickstart
new developments in AI that would in turn inform and improve the development
of systems focused on military problems.  Also, violence is not limited to
nation-state level conflict.  Some of the challenges outlined herein may
ultimately arise in civilian contexts.

For industry, the incentives for developing adaptable AI are currently low,
but here is an opportunity for the defense community to once more contribute
to technological advances in ways that benefit the broader population, much in
the same way that the development of the B-52 \enquote{spilled over} into the
development of the Boeing 707 and subsequent commercial jet
aircraft~\cite{kotha2010spillovers}.

Ethically-aware AI support systems would be useful across many human domains,
including law, diplomacy, and negotiations between multiple parties.  As war
has been often described as the application of force as a replacement for
diplomacy, perhaps AI-enhanced diplomacy can reduce the risk of a
hyper-accelerated AI war.

Finally, this paper assumes that AI does not ultimately \enquote{run away}
from human ability to control it.  Serious philosophers view this sort of
\enquote{technological singularity} as a realistic scenario, and we believe it
must be addressed no later than widespread use of human-on-the-loop systems.

This paper presents a framework for understanding the integration of AI and ML
into military weapons systems.  The steps we take may be on a path to human
obsolescence as combatants, and the decision to proceed on this path should be
well informed and involve all members of our societies.  Given that other
players are deeply engaged in the weaponization of AI, we would be foolish not
to research and experiment with the development of highly capable systems.
But we believe that a better answer may be to to support regulation and
prohibition because, like chemical and biological weapons, for weaponized AI,
\enquote{the only winning move is not to play.}



\end{document}